\PassOptionsToPackage{unicode}{hyperref}
\PassOptionsToPackage{hyphens}{url}
\documentclass[
]{article}
\usepackage{xcolor}
\usepackage{amsmath,amssymb}
\usepackage{iftex}
\ifPDFTeX
  \usepackage[T1]{fontenc}
  \usepackage[utf8]{inputenc}
  \usepackage{textcomp} % provide euro and other symbols
\else % if luatex or xetex
  \usepackage{unicode-math} % this also loads fontspec
  \defaultfontfeatures{Scale=MatchLowercase}
  \defaultfontfeatures[\rmfamily]{Ligatures=TeX,Scale=1}
\fi
\usepackage{lmodern}
\ifPDFTeX\else
\fi
\IfFileExists{upquote.sty}{\usepackage{upquote}}{}
\IfFileExists{microtype.sty}{% use microtype if available
  \usepackage[]{microtype}
  \UseMicrotypeSet[protrusion]{basicmath} % disable protrusion for tt fonts
}{}
\makeatletter
\@ifundefined{KOMAClassName}{% if non-KOMA class
  \IfFileExists{parskip.sty}{%
    \usepackage{parskip}
  }{% else
    \setlength{\parindent}{0pt}
    \setlength{\parskip}{6pt plus 2pt minus 1pt}}
}{% if KOMA class
  \KOMAoptions{parskip=half}}
\makeatother
\usepackage{color}
\usepackage{fancyvrb}

\DefineVerbatimEnvironment{Highlighting}{Verbatim}{commandchars=\\\{\}}
\newenvironment{Shaded}{}{}

\newcommand{\BuiltInTok}[1]{\textcolor[rgb]{0.00,0.50,0.00}{#1}}

\newcommand{\CommentTok}[1]{\textcolor[rgb]{0.38,0.63,0.69}{\textit{#1}}}

\newcommand{\ControlFlowTok}[1]{\textcolor[rgb]{0.00,0.44,0.13}{\textbf{#1}}}

\newcommand{\FunctionTok}[1]{\textcolor[rgb]{0.02,0.16,0.49}{#1}}

\newcommand{\KeywordTok}[1]{\textcolor[rgb]{0.00,0.44,0.13}{\textbf{#1}}}
\newcommand{\NormalTok}[1]{#1}
\newcommand{\OperatorTok}[1]{\textcolor[rgb]{0.40,0.40,0.40}{#1}}

\newcommand{\SpecialCharTok}[1]{\textcolor[rgb]{0.25,0.44,0.63}{#1}}
\newcommand{\SpecialStringTok}[1]{\textcolor[rgb]{0.73,0.40,0.53}{#1}}
\newcommand{\StringTok}[1]{\textcolor[rgb]{0.25,0.44,0.63}{#1}}
\newcommand{\VariableTok}[1]{\textcolor[rgb]{0.10,0.09,0.49}{#1}}

\NewDocumentCommand\citeproctext{}{}
\NewDocumentCommand\citeproc{mm}{%
  \begingroup\def\citeproctext{#2}\cite{#1}\endgroup}
\makeatletter
 \let\@cite@ofmt\@firstofone
 \def\@biblabel#1{}
 \def\@cite#1#2{{#1\if@tempswa , #2\fi}}
\makeatother
\newlength{\cslhangindent}
\newlength{\csllabelwidth}
\newenvironment{CSLReferences}[2] % #1 hanging-indent, #2 entry-spacing
 {\begin{list}{}{%
  \setlength{\itemindent}{0pt}
  \setlength{\leftmargin}{0pt}
  \setlength{\parsep}{0pt}
  \ifodd #1
   \setlength{\leftmargin}{\cslhangindent}
   \setlength{\itemindent}{-1\cslhangindent}
  \fi
  \setlength{\itemsep}{#2\baselineskip}}}
 {\end{list}}
\usepackage{calc}

\newcommand{\CSLLeftMargin}[1]{\parbox[t]{\csllabelwidth}{\strut#1\strut}}
\newcommand{\CSLRightInline}[1]{\parbox[t]{\linewidth - \csllabelwidth}{\strut#1\strut}}

\providecommand{\tightlist}{%
  \setlength{\itemsep}{0pt}\setlength{\parskip}{0pt}}
\usepackage{bookmark}
\IfFileExists{xurl.sty}{\usepackage{xurl}}{} % add URL line breaks if available
\hypersetup{
  pdftitle={Semantic Laundering in AI Agent Architectures: Why Tool Boundaries Do Not Confer Epistemic Warrant},
  pdfauthor={Oleg Romanchuk; Roman Bondar},
  hidelinks,
  pdfcreator={LaTeX via pandoc}}

\title{Semantic Laundering in AI Agent Architectures: Why Tool
Boundaries Do Not Confer Epistemic Warrant}
\author{Oleg Romanchuk \and Roman Bondar}
\date{}

\begin{document}
\maketitle

\subsection{Abstract}\label{abstract}

LLM-based agent architectures systematically conflate information
transport mechanisms with epistemic justification mechanisms. We
formalize this class of architectural failures as \textbf{semantic
laundering} --- a pattern where propositions with absent or weak warrant
are accepted by the system as admissible by crossing architecturally
trusted interfaces.

We show that semantic laundering constitutes an \textbf{architectural
realization of the Gettier problem}: propositions acquire high epistemic
status without a connection between their justification and what makes
them true. Unlike classical Gettier cases, this effect is not accidental
--- it is architecturally determined and systematically reproducible.

The central result is the \textbf{Theorem of Inevitable Self-Licensing}:
under standard architectural assumptions, circular epistemic
justification cannot be eliminated. We introduce the \textbf{Warrant
Erosion Principle} as the fundamental explanation for this effect and
show that scaling, model improvement, and LLM-as-judge schemes are
structurally incapable of eliminating a problem that exists at the type
level.

\subsection{1. Introduction}\label{introduction}

Agent frameworks
\citeproc{ref-yao2023ReActSynergizingReasoning}{{[}1{]}},
\citeproc{ref-chase2022LangChain}{{[}2{]}},
\citeproc{ref-2026NVlabsToolOrchestra}{{[}3{]}} provide language models
with the capability to interact with external tools. The common
architectural pattern: an agent calls a tool, receives a result, and
uses it as a basis for subsequent decisions.

We identify a structural defect: these architectures conflate
information transport mechanisms (how information arrived) with
epistemic justification (why it should be believed), as manifested in
common agent implementations \citeproc{ref-langchainAgents}{{[}4{]}}. An
LLM ``expert'' wrapped in a tool receives the same epistemic status as a
database query. The tool boundary itself functions as an epistemic
laundering mechanism.

\textbf{Semantic laundering} is an architectural pattern where a weakly
warranted proposition, by passing through a trusted interface, is
accepted by the system as an observation with high epistemic status
without epistemically relevant inference.

This is not an accidental implementation error, but a structural
architectural limitation that makes fundamentally unattainable, at the
architectural level:

\begin{itemize}
\tightlist
\item
  Elimination of begging the question in adopted decisions (petitio
  principii, regulatory compliance),
\item
  Restoration of logically correct connection between decisions and
  observations (non sequitur, auditability),
\item
  Formal limitation of epistemic justification quality (idem per idem,
  safety guarantees).
\end{itemize}

\subsubsection{1.1 Contributions}\label{contributions}

\begin{enumerate}
\def\labelenumi{\arabic{enumi}.}
\item
  \textbf{Formalization of semantic laundering} as an architectural
  realization of the Gettier problem --- a failure class distinct from
  hallucinations, prompt injections, and reward hacking.
\item
  \textbf{Warrant Erosion Principle}: any justification that passes
  through an interpreting process may lose part of its epistemic force,
  since the connection to truth-makers is not preserved.
\item
  \textbf{Theorem of Inevitable Self-Licensing}: under common
  architectural assumptions, circular justification of decisions is
  inevitable in the absence of explicit epistemic tool typing.
\item
  \textbf{Corollary for LLM-as-judge}: evaluation is structurally
  unsound when judge and agent operate in the same assumption space.
\item
  \textbf{Architectural direction}: epistemic typing of tools at the
  type level, explaining divergence between channel-based and
  content-based warrant under identical content.
\end{enumerate}

\subsubsection{1.2 Scope and Positioning}\label{scope-and-positioning}

This work is \textbf{positional and architectural in nature} and belongs
to the field of architectural epistemology --- the study of how AI
systems represent, transmit, and use epistemic warrant for their
propositions. We analyze not the quality of models or empirical metrics,
but \textbf{structural conditions} under which epistemically correct
justification is fundamentally impossible.

\textbf{The central thesis} is that typical agent architectures enable
semantic laundering \textbf{regardless of model quality, training data,
and calibration schemes}, since the source of this failure class lies in
architecture, not in generative properties of components.

\textbf{Philosophical position}: We deliberately adopt an
\textbf{anti-internalist position} regarding epistemic justification.
Internal coherence, argument quality, confidence, or other
characteristics of the generative process are not considered sufficient
grounds for knowledge. The epistemic status of a proposition is
determined exclusively by its connection to observations and admissible
inference rules, not by properties or quality of the generating
component. This assumption is used as a minimal architectural
requirement and does not depend on accepting any complete
epistemological theory.

\subsection{2. Background and Related
Work}\label{background-and-related-work}

\subsubsection{2.1 Agent Architectures}\label{agent-architectures}

\textbf{ReAct} \citeproc{ref-yao2023ReActSynergizingReasoning}{{[}1{]}}
--- the standard agent architectural pattern where reasoning and acting
stages alternate: the agent generates reasoning, executes an action
(tool call), receives the result as an observation, and continues the
cycle.

\textbf{ToolOrchestra} \citeproc{ref-2026NVlabsToolOrchestra}{{[}3{]}}
--- an RL-oriented framework for tool orchestration in agent workflows,
where tools are typed by functional signature and interface, but not by
their epistemic class.

\textbf{LangChain} \citeproc{ref-chase2022LangChain}{{[}2{]}} --- an
orchestration framework that actually implements composition of
specialized LLM components; no architectural distinction is made between
hypothesis generation (LLM calls) and observation acquisition (external
APIs or tool calls)

\textbf{Common pattern}: tool outputs → observations → ground subsequent
reasoning.

Tool call results are treated as observations without distinction
between:

\begin{itemize}
\tightlist
\item
  Tools that observe world state (DB queries, sensors),
\item
  Tools that generate propositions (LLM experts).
\end{itemize}

In the limiting case, when tool outputs are uniformly treated as
observations, such frameworks implement not agent interaction with the
world, but closed orchestration of LLM experts, where epistemic
justification arises exclusively from internal composition of generative
components.

\subsubsection{2.2 Epistemic Foundations}\label{epistemic-foundations}

Classical epistemology treats knowledge as a combination of belief,
truth, and justification. In the \textbf{Justified True Belief}
formulation \citeproc{ref-gettier1963JustifiedTrueBelief}{{[}5{]}},
knowledge is defined as a true belief possessing proper justification.
The key element here is not the belief itself or its truth, but the
character of the justification connecting the belief to the
corresponding fact.

In \textbf{warrant} theories
\citeproc{ref-plantinga1993WarrantCurrentDebate}{{[}6{]}}, what is
considered is not merely justified belief, but what makes a true belief
knowledge --- that is, what characteristic of a true belief makes it
epistemically admissible as knowledge. Plantinga critiques common
approaches to justification and emphasizes that knowledge requires a
deeper epistemic criterion than mere coincidence of truth, belief, and
accidental causal connections.

\textbf{Our contribution} is that we transfer these epistemological
distinctions to the AI agent architecture level and show that common
agent frameworks systematically ignore them, erasing the boundary
between observation and generation. Semantic laundering is an
\textbf{architectural form of the Gettier problem}: propositions receive
knowledge status without a connection between justification and
truth-maker.

\subsubsection{2.3 LLM-as-Judge}\label{llm-as-judge}

Recent work explores using language models as generation quality
evaluators (LLM-as-judge). Zheng et al.
\citeproc{ref-zheng2023JudgingLLMasaJudgeMTBench}{{[}7{]}} document
systematic biases in such evaluators, including position and verbosity
bias, as well as self-enhancement tendency and limited reasoning
capability, affecting evaluation reliability. Liu et al.
\citeproc{ref-liu2023GEvalNLGEvaluation}{{[}8{]}} in the G-Eval
framework also emphasize that LLM-based evaluators do not always
correlate with human judgments and demonstrate biases in evaluations.

We argue that these effects are not merely empirical noise, but symptoms
of a deeper architectural problem: in LLM-as-judge schemes, evaluation
and generation are performed in the same epistemic space, so that the
model's judgment is used to justify judgments of the same nature,
leading to circular justification that cannot be eliminated by improving
the model or calibrating the evaluator. In other words, evaluation and
generation operate over the same epistemic type, making warrant circular
by construction.

\subsubsection{2.4 Safety-Critical AI}\label{safety-critical-ai}

FDA guidance
\citeproc{ref-u.s.foodanddrugadministration2021ArtificialIntelligenceMachine}{{[}9{]}}
on regulating AI/ML-Based Software as a Medical Device emphasizes the
need for data transparency, algorithm logic, and modification plans,
provides for collection and analysis of real-world data, and control of
changes throughout the product lifecycle. These requirements reflect a
general class of requirements for safety-critical systems, where safety,
traceability of decisions to underlying data and algorithmic logic, and
justification verification are key conditions for operational approval.
In particular, regulatory guidance requires that system decisions be
traceable to identifiable data sources and algorithmic transformations,
rather than opaque internal model states or undocumented heuristics
\citeproc{ref-u.s.foodanddrugadministration2021ArtificialIntelligenceMachine}{{[}9{]}}.

Our thesis is that architectures permitting semantic laundering are
structurally incompatible with such requirements. In such systems, it is
impossible to show that a decision relies on observations rather than
self-confirming generative inferences; it is impossible to reconstruct
the justification chain; it is impossible to formally specify admissible
grounds for making decisions.

\subsection{3. Formalization}\label{formalization}

\subsubsection{3.1 Epistemic Primitives}\label{epistemic-primitives}

\textbf{Definition 1 (epistemic warrant)}: A proposition \(P\) has
epistemic warrant \(W\) if there exists a justification chain linking
\(P\) to observations \(O\) through admissible inference rules \(I\).

\[
\begin{aligned}
\text{warrant}(P) &= \langle O, I \rangle \\
\text{where } O &= \{\text{observations grounding } P\} \\
              I &= \{\text{inference rules deriving } P \text{ from } O\}
\end{aligned}
\]

\textbf{Note on warrant strength}: We use the terms ``weak'' and
``strong'' as \textbf{partial order labels} over warrant states. A
warrant state is considered \emph{stronger} than another if it is
strictly greater under inclusion of observations or admissible inference
rules, i.e., if it is grounded in a superset of observations and/or a
strictly stronger set of validated inference rules.

\textbf{Definition 2 (epistemic relevance)}: An inference step I is
\textbf{epistemically relevant} if and only if it increases warrant(P)
by:

\begin{itemize}
\tightlist
\item
  \begin{enumerate}
  \def\labelenumi{(\alph{enumi})}
  \tightlist
  \item
    introducing new observations \(O' \notin O\), OR
  \end{enumerate}
\item
  \begin{enumerate}
  \def\labelenumi{(\alph{enumi})}
  \setcounter{enumi}{1}
  \tightlist
  \item
    introducing new validated inference rules \(R' \notin R\)
  \end{enumerate}
\end{itemize}

\textbf{Note on validated rules}: Validated inference rules are rules
whose correctness is externally established (mathematical derivations,
verified algorithms, domain axioms), not stochastic generation.

\textbf{Corollary}: Crossing a tool interface is not epistemically
relevant if it satisfies neither condition (a) nor (b).

\textbf{Definition 3 (semantic laundering)}: An architecture exhibits
semantic laundering if:

\begin{enumerate}
\def\labelenumi{\arabic{enumi}.}
\tightlist
\item
  There exists a proposition \(P_1\) with weak warrant \(W_1\);
\item
  \(P_1\) passes through a boundary \(B\) (tool call, interface, model);
\item
  The output is a proposition \(P_2\) with strong warrant \(W_2\);
\item
  No epistemically relevant inference step occurs between \(P_1\) and
  \(P_2\).
\end{enumerate}

\[
\begin{aligned}
\text{SL}(B) = \exists P_1, P_2: &\text{warrant}(P_1) = \text{weak} \land \\
                                 &P_2 = B(P_1) \land \\
                                 &\text{warrant}(P_2) = \text{strong} \land \\
                                 &\text{epistemic\_inferences}(P_2) = \text{epistemic\_inferences}(P_1)
\end{aligned}
\]

\subsubsection{3.2 Causal vs Epistemic
Transitions}\label{causal-vs-epistemic-transitions}

\textbf{Definition 4 (causal transition)}: An operation that changes
system state.

\textbf{Definition 5 (epistemic transition)}: An operation that
increases epistemic warrant of a proposition.

\textbf{Key distinction}: A tool call is causal by default and
epistemically inert unless typed otherwise. Causal relevance (the tool
changes system state) does not entail epistemic relevance (an increase
in epistemic warrant).

\[
\text{causal\_inferences}(P_2) \neq \text{epistemic\_inferences}(P_2)
\]

In particular, a tool boundary may causally transform \(P_1\) into
\(P_2\) without providing any epistemic grounding for \(P_2\).

\subsubsection{3.3 Warrant Erosion
Principle}\label{warrant-erosion-principle}

\textbf{Definition 6 (warrant erosion)}: Interpreting or generative
operations do not automatically preserve epistemic warrant, since they
may break the connection between a proposition and its truth-maker.

\textbf{Clarification}: The Warrant Erosion Principle does not claim
that all interpretation destroys knowledge or makes it impossible. It
claims a weaker and architecturally significant position:
\textbf{epistemic warrant is not automatically preserved through
interpreting or generative transformations} unless the architecture
specifies explicit epistemic guarantees for preserving the connection
between proposition and truth-maker.

\textbf{Justification}: Any epistemic access available to an agent is
mediated --- through cognitive, instrumental, or formal processes. This
mediation means that the connection between a proposition and what makes
it true is not automatically preserved through transformations; similar
diagnostic concerns have been articulated in the Epistemic Suite
\citeproc{ref-kelly2025EpistemicSuitePostFoundational}{{[}10{]}}, which
surfaces confidence laundering and other epistemic distortions in AI
outputs.

\textbf{Corollary 1}: Gettier situations are not anomalies, but
structural invariants of systems that generate epistemic claims through
mediated processes without explicit warrant preservation.

\textbf{Corollary 2}: Improving generation quality (more coherent
argumentation, better approximation) is not epistemically relevant if
the observation set \(O\) has not changed. Quality of generation
\(\neq\) quality of warrant.

\textbf{Connection to Gettier}: Warrant erosion explains this gap: the
connection to truth-makers is not preserved by default once
justification passes through interpreting processes.

\subsubsection{3.4 Theorem: Inevitable
Self-Licensing}\label{theorem-inevitable-self-licensing}

\textbf{Theorem 1 (Inevitable Self-Licensing)}

\textbf{Note on scope}: This is an impossibility result under given
architectural assumptions. The assumptions formulated below are not
logically necessary, but they correspond to the dominant architectural
pattern of contemporary agent frameworks
(\citeproc{ref-yao2023ReActSynergizingReasoning}{{[}1{]}},
\citeproc{ref-chase2022LangChain}{{[}2{]}},
\citeproc{ref-2026NVlabsToolOrchestra}{{[}3{]}}, and similar). The
theorem asserts that within this architectural paradigm, circular
epistemic justification cannot be eliminated.

\textbf{Assumptions}:

\begin{enumerate}
\def\labelenumi{\arabic{enumi}.}
\tightlist
\item
  Agent and tool outputs belong to the same proposition type \(P\) and
  are treated by the architecture as interchangeable. Intuitively, this
  means that agent and tool outputs are processed by the architecture as
  elements of the same proposition space, without distinction by
  epistemic role.
\item
  Tool call results are accepted as observations \(O \subset P\)
  \textbf{without regard to the epistemic type of the tool} (see §3.5)
\item
  The architecture assigns epistemic status to propositions via a
  function
  \(A: P \to \{\text{ASSERTIVE}, \text{CONDITIONAL}, \text{REFUSAL}\}\),
  and LLM-generated propositions may influence this assignment.
\end{enumerate}

\textbf{Claim}: Under these assumptions, circular epistemic
justification is inevitable.

\textbf{Limitation of the result}: The theorem does not claim that
self-licensing is logically necessary for all agent architectures, nor
does it claim a universal ontological conclusion. It establishes the
impossibility of eliminating circular epistemic justification within the
dominant architectural pattern, where epistemic roles of tools are not
explicitly distinguished.

\textbf{Proof sketch}:

Let \(\text{LLM}_1\) be the agent, \(\text{LLM}_2\) be an ``expert''
tool.

\begin{enumerate}
\def\labelenumi{\arabic{enumi}.}
\tightlist
\item
  \(\text{LLM}_1\) generates proposition \(p_1 \in P\)
\item
  Tool \(T\) wraps \(\text{LLM}_2\) call: \(p_2 = T(\text{LLM}_2(p_1))\)
\item
  By Assumption 2: \(p_2\) is accepted as observation, i.e.,
  \(p_2 \in O\)
\item
  By Assumption 3: \(p_2\) is used as input to the epistemic status
  function \(A(p_1)\)
\item
  Since \(p_2 \in P\), the status of \(p_1\) is determined by a
  proposition of the same type
\item
  Therefore, \(A(p_1)\) is influenced by an element of \(P\) produced by
  the same epistemic process as \(p_1\) itself
\end{enumerate}

Circular epistemic justification arises: propositions participate in
assigning their own epistemic status via architectural mediation.
\(\blacksquare\)

\textbf{Corollary 3}: Evaluation in LLM-as-judge schemes is structurally
unsound if the evaluator and the evaluated agent operate in the same
proposition space without epistemic typing.

\textbf{Corollary 4}: Increasing model capability (e.g., replacing
\(\text{LLM}_2\) with a stronger one) does not eliminate self-licensing:
the problem is architectural, not generative.

\subsubsection{3.5 Tool Epistemic Classes}\label{tool-epistemic-classes}

The critical weakness of Assumption 2 is that all tools are treated by
the architecture as homogeneous. In practice, tools differ fundamentally
in their epistemic character.

\textbf{Definition 7 (tool epistemic class)}: A tool's epistemic class
determines whether its results can legitimately be treated as
observations.

\begin{verbatim}
EpistemicClass :=
  OBSERVER      // observations (DB queries, APIs, sensors)
  COMPUTATION   // computed results (math, verified algorithms)
  GENERATOR     // propositions (LLM, stochastic generation)
\end{verbatim}

OBSERVER: External world state → can serve as observation (e.g.,
database queries, sensors)

COMPUTATION: Derived from observations → can serve as observation (e.g.,
sorting, filtering)

GENERATOR: Propositions from \(P\) → \textbf{cannot serve as
observation} (e.g., LLM outputs)

COMPUTATION tools do not introduce new propositions about world state,
but only deterministically transform existing observations. By contrast,
GENERATOR tools necessarily introduce mediated propositions whose
warrant is subject to erosion unless explicitly preserved by
architectural constraints.

\textbf{Significance for Theorem 1}:

Assumption 2 becomes problematic \textbf{only} when GENERATOR tools are
admitted to the observation set. OBSERVER tools truly produce
observations --- a database query returns actual external world state.

The architectural blindness is precisely this: \textbf{common frameworks
do not distinguish epistemic classes of tools}. All tools are processed
uniformly:

\begin{Shaded}
\begin{Highlighting}[]
\CommentTok{\# Typical pattern — absence of epistemic distinction}
\KeywordTok{def}\NormalTok{ execute\_tool(tool, args):}
\NormalTok{    result }\OperatorTok{=}\NormalTok{ tool.call(args)}
    \ControlFlowTok{return}\NormalTok{ Observation(result)  }\CommentTok{\# ← any result is treated }
                                \CommentTok{\#   as observation}
\end{Highlighting}
\end{Shaded}

This is precisely where semantic laundering arises:

\begin{Shaded}
\begin{Highlighting}[]
\CommentTok{\# LLM{-}based "expert" tool}
\NormalTok{expert\_tool }\OperatorTok{=}\NormalTok{ Tool(}
\NormalTok{    name}\OperatorTok{=}\StringTok{"risk\_assessor"}\NormalTok{,}
\NormalTok{    function}\OperatorTok{=}\KeywordTok{lambda}\NormalTok{ x: LLM.call(}\SpecialStringTok{f"Assess risk: }\SpecialCharTok{\{}\NormalTok{x}\SpecialCharTok{\}}\SpecialStringTok{"}\NormalTok{),}
    \CommentTok{\# epistemic class not specified (key problem)}
\NormalTok{)}

\NormalTok{result }\OperatorTok{=}\NormalTok{ expert\_tool.call(}\StringTok{"Task X"}\NormalTok{)}
\CommentTok{\# result: "High risk" (LLM hypothesis)}

\NormalTok{context.add\_observation(result)  }\CommentTok{\# ← GENERATOR result treated }
                                 \CommentTok{\#   as OBSERVER}
\end{Highlighting}
\end{Shaded}

\textbf{Corollary 5}: Semantic laundering is structurally prevented in
architectures that explicitly type GENERATOR tools and exclude their
results from the observation set \(O\).

\subsection{4. Architectural
Instantiation}\label{architectural-instantiation}

\textbf{Note}: This section demonstrates concrete architectural
realizations of Theorem 1. The theorem is a logical consequence of its
premises and does not require empirical validation; the examples shown
demonstrate how the corresponding failure inevitably arises in common
agent runtimes under standard architectural decisions.

\subsubsection{4.1 ReAct Expert Tool
Pattern}\label{react-expert-tool-pattern}

The most typical manifestation of Theorem 1 occurs in ReAct-style
architectures where so-called \emph{expert tools} are wrappers over LLM
calls.

The key point is not the use of tools itself, but \textbf{how the agent
runtime types the result of their execution}.

Below is a minimal but architecturally correct model of such a runtime.

\begin{Shaded}
\begin{Highlighting}[]
\CommentTok{\# =========================}
\CommentTok{\# Tool Contract}
\CommentTok{\# =========================}

\KeywordTok{class}\NormalTok{ Tool:}
    \KeywordTok{def} \FunctionTok{\_\_init\_\_}\NormalTok{(}\VariableTok{self}\NormalTok{, name, description, function):}
        \VariableTok{self}\NormalTok{.name }\OperatorTok{=}\NormalTok{ name}
        \VariableTok{self}\NormalTok{.description }\OperatorTok{=}\NormalTok{ description}
        \VariableTok{self}\NormalTok{.function }\OperatorTok{=}\NormalTok{ function}


\KeywordTok{class}\NormalTok{ RiskExpertTool(Tool):}
    \KeywordTok{def} \FunctionTok{\_\_init\_\_}\NormalTok{(}\VariableTok{self}\NormalTok{, llm):}
        \BuiltInTok{super}\NormalTok{().}\FunctionTok{\_\_init\_\_}\NormalTok{(}
\NormalTok{            name}\OperatorTok{=}\StringTok{"risk\_expert"}\NormalTok{,}
\NormalTok{            description}\OperatorTok{=}\StringTok{"Assesses task risk"}\NormalTok{,}
\NormalTok{            function}\OperatorTok{=}\VariableTok{self}\NormalTok{.\_call}
\NormalTok{        )}
        \VariableTok{self}\NormalTok{.llm }\OperatorTok{=}\NormalTok{ llm}

    \KeywordTok{def}\NormalTok{ \_call(}\VariableTok{self}\NormalTok{, task: }\BuiltInTok{str}\NormalTok{) }\OperatorTok{{-}\textgreater{}} \BuiltInTok{str}\NormalTok{:}
        \CommentTok{\# LLM{-}based GENERATOR tool}
        \ControlFlowTok{return} \VariableTok{self}\NormalTok{.llm.call(}\SpecialStringTok{f"Assess task risk: }\SpecialCharTok{\{}\NormalTok{task}\SpecialCharTok{\}}\SpecialStringTok{"}\NormalTok{)}


\CommentTok{\# =========================}
\CommentTok{\# Tool Registry (Runtime Level)}
\CommentTok{\# =========================}

\NormalTok{llm\_agent }\OperatorTok{=}\NormalTok{ LLM(}\StringTok{"gpt{-}5{-}mini"}\NormalTok{)}
\NormalTok{llm\_expert }\OperatorTok{=}\NormalTok{ LLM(}\StringTok{"gpt{-}5.2"}\NormalTok{)}

\NormalTok{tool\_registry }\OperatorTok{=}\NormalTok{ \{}
    \StringTok{"risk\_expert"}\NormalTok{: RiskExpertTool(llm\_expert)}
\NormalTok{\}}

\CommentTok{\# Tool descriptions passed to LLM}
\NormalTok{tool\_specs }\OperatorTok{=}\NormalTok{ [}
\NormalTok{    \{}
        \StringTok{"name"}\NormalTok{: tool.name,}
        \StringTok{"description"}\NormalTok{: tool.description,}
\NormalTok{    \}}
    \ControlFlowTok{for}\NormalTok{ tool }\KeywordTok{in}\NormalTok{ tool\_registry.values()}
\NormalTok{]}


\CommentTok{\# =========================}
\CommentTok{\# Agent Runtime (ReAct Style)}
\CommentTok{\# =========================}

\KeywordTok{class}\NormalTok{ Agent:}
    \KeywordTok{def} \FunctionTok{\_\_init\_\_}\NormalTok{(}\VariableTok{self}\NormalTok{, llm, tools):}
        \VariableTok{self}\NormalTok{.llm }\OperatorTok{=}\NormalTok{ llm}
        \VariableTok{self}\NormalTok{.tools }\OperatorTok{=}\NormalTok{ tools}
        \VariableTok{self}\NormalTok{.context }\OperatorTok{=}\NormalTok{ []}

    \KeywordTok{def}\NormalTok{ reason(}\VariableTok{self}\NormalTok{, task: }\BuiltInTok{str}\NormalTok{) }\OperatorTok{{-}\textgreater{}} \BuiltInTok{str}\NormalTok{:}
        \CommentTok{\# Step 1: LLM generates reasoning and may }
        \CommentTok{\#         initiate tool\_call}
\NormalTok{        user\_msg }\OperatorTok{=}\NormalTok{ \{}\StringTok{"role"}\NormalTok{: }\StringTok{"user"}\NormalTok{, }\StringTok{"content"}\NormalTok{: }\SpecialStringTok{f"Task: }\SpecialCharTok{\{}\NormalTok{task}\SpecialCharTok{\}}\SpecialStringTok{"}\NormalTok{\}}
        \VariableTok{self}\NormalTok{.context.append(user\_msg)}
        
        \CommentTok{\# LLM decides to call risk\_expert tool}
\NormalTok{        msg }\OperatorTok{=} \VariableTok{self}\NormalTok{.llm.call(}
\NormalTok{            messages}\OperatorTok{=}\VariableTok{self}\NormalTok{.context,}
\NormalTok{            tools}\OperatorTok{=}\NormalTok{tool\_specs}
\NormalTok{        )}
        
        \CommentTok{\# Add assistant response to context}
        \VariableTok{self}\NormalTok{.context.append(msg)}

        \CommentTok{\# Step 2: runtime executes tool call}
        \ControlFlowTok{if}\NormalTok{ msg.get(}\StringTok{"tool\_call"}\NormalTok{):}
\NormalTok{            tool\_name }\OperatorTok{=}\NormalTok{ msg[}\StringTok{"tool\_call"}\NormalTok{][}\StringTok{"name"}\NormalTok{]}
\NormalTok{            args }\OperatorTok{=}\NormalTok{ msg[}\StringTok{"tool\_call"}\NormalTok{][}\StringTok{"arguments"}\NormalTok{]}

\NormalTok{            tool }\OperatorTok{=} \VariableTok{self}\NormalTok{.tools[tool\_name]}
\NormalTok{            result }\OperatorTok{=}\NormalTok{ tool.function(}\OperatorTok{**}\NormalTok{args)}

            \CommentTok{\# SEMANTIC LAUNDERING OCCURS HERE}
            \VariableTok{self}\NormalTok{.context.append(\{}
                \StringTok{"role"}\NormalTok{: }\StringTok{"tool"}\NormalTok{,}
                \StringTok{"content"}\NormalTok{: result}
\NormalTok{            \})}

        \CommentTok{\# Step 3: LLM uses "tool result" as epistemic basis}
\NormalTok{        final\_user\_msg }\OperatorTok{=}\NormalTok{ \{}\StringTok{"role"}\NormalTok{: }\StringTok{"user"}\NormalTok{, }\StringTok{"content"}\NormalTok{: }\StringTok{"Suggest action"}\NormalTok{\}}
        \VariableTok{self}\NormalTok{.context.append(final\_user\_msg)}
        
\NormalTok{        final }\OperatorTok{=} \VariableTok{self}\NormalTok{.llm.call(messages}\OperatorTok{=}\VariableTok{self}\NormalTok{.context)}
        \VariableTok{self}\NormalTok{.context.append(final)}

        \ControlFlowTok{return}\NormalTok{ final}
\end{Highlighting}
\end{Shaded}

\textbf{Architectural trace}:

\begin{enumerate}
\def\labelenumi{\arabic{enumi}.}
\tightlist
\item
  Agent (\(\text{LLM}_1\)) generates reasoning
\item
  Agent initiates tool call \texttt{risk\_expert}
\item
  Tool executes and calls \(\text{LLM}_2\) → returns \(p_2\)
\item
  Runtime types \(p_2\) as tool result (treated as observation) and adds
  to context
\item
  \(\text{LLM}_1\) uses \(p_2\) as epistemic basis
\item
  Final recommendation is generated
\end{enumerate}

However:

\begin{itemize}
\tightlist
\item
  \(p_2\) is a proposition, not an observation (\(p_2 \in P\))
\item
  \(p_2\) is generated by a language model (\(\text{LLM}_2\)),
  epistemically equivalent to \(\text{LLM}_1\)
\item
  No new observations \(O'\) are introduced
\item
  No validated inference rules are applied
\end{itemize}

Thus, the apparent epistemic justification arises exclusively from the
runtime's architectural decision to type any tool result as an
observation.

\textbf{This is an instantiation of Theorem 1}:
\(\text{LLM}_1 \to \text{Tool}(\text{LLM}_2) \to\) epistemic status
elevation of \(\text{LLM}_1\). The agent validates itself through
architectural indirection.

\subsubsection{4.2 Multi-Agent Validation
Pattern}\label{multi-agent-validation-pattern}

The second common architectural pattern where Theorem 1 manifests is
multi-agent validation schemes.

\begin{Shaded}
\begin{Highlighting}[]
\CommentTok{\# Agent 1 formulates proposition}
\NormalTok{agent1 }\OperatorTok{=}\NormalTok{ Agent(}\StringTok{"analyst"}\NormalTok{)}
\NormalTok{claim }\OperatorTok{=}\NormalTok{ agent1.analyze(task)}
\CommentTok{\# claim: "Task X should be prioritized"}

\CommentTok{\# Agent 2 "validates" proposition (also LLM)}
\NormalTok{agent2 }\OperatorTok{=}\NormalTok{ Agent(}\StringTok{"validator"}\NormalTok{)}
\NormalTok{validation }\OperatorTok{=}\NormalTok{ agent2.validate(claim)}
\CommentTok{\# validation: "Analysis is correct, recommendation is justified"}

\CommentTok{\# System treats validation as evidence}
\NormalTok{claim.evidence.add(validation)  }\CommentTok{\# ← self{-}licensing}
\NormalTok{claim.status }\OperatorTok{=} \StringTok{"validated"}      \CommentTok{\# ← laundering complete}
\end{Highlighting}
\end{Shaded}

\textbf{Why this is self-licensing}:

\begin{itemize}
\tightlist
\item
  \(\text{claim} \in P\) (proposition space)
\item
  \(\text{validation} \in P\) (same space)
\item
  validation generated by LLM (same epistemic class)
\item
  agent1 → agent2.validate() → upgrades claim status
\end{itemize}

No observations introduced. No external confirmation. Proposition
validates proposition.

\subsubsection{4.3 Channel-Based vs Content-Based
Warrant}\label{channel-based-vs-content-based-warrant}

Both patterns are based on the same mechanism.

\begin{Shaded}
\begin{Highlighting}[]
\CommentTok{\# WRONG: Channel{-}based warrant}
\KeywordTok{def}\NormalTok{ derive\_strength(node):}
    \ControlFlowTok{if}\NormalTok{ node.source }\OperatorTok{==} \StringTok{"tool"}\NormalTok{:}
        \ControlFlowTok{return} \StringTok{"strong"}   \CommentTok{\# ← all tools trusted equally}
    \ControlFlowTok{else}\NormalTok{:}
        \ControlFlowTok{return} \StringTok{"weak"}
\end{Highlighting}
\end{Shaded}

\textbf{Consequence}: LLM tool output gets ``strong'' warrant despite
being a hypothesis with unknown confidence.

Comparison with content-based warrant (laundering-resistant):

\begin{Shaded}
\begin{Highlighting}[]
\CommentTok{\# CORRECT: Content{-}based warrant}
\KeywordTok{def}\NormalTok{ derive\_strength(node):}
    \ControlFlowTok{if}\NormalTok{ node.epistemic\_class }\OperatorTok{==} \StringTok{"OBSERVER"}\NormalTok{:}
        \ControlFlowTok{if}\NormalTok{ node.status }\OperatorTok{==} \StringTok{"confirmed"}\NormalTok{:}
            \ControlFlowTok{return} \StringTok{"strong"}
    \ControlFlowTok{elif}\NormalTok{ node.epistemic\_class }\OperatorTok{==} \StringTok{"GENERATOR"}\NormalTok{:}
        \ControlFlowTok{return} \StringTok{"weak"}  \CommentTok{\# always weak, regardless of channel}
    \ControlFlowTok{return} \StringTok{"weak"}
\end{Highlighting}
\end{Shaded}

\subsubsection{4.4 Interpretation}\label{interpretation}

The examples demonstrate:

\begin{enumerate}
\def\labelenumi{\arabic{enumi}.}
\item
  \textbf{Architectural blindness}: Without epistemic class distinction
  (§3.5), systems cannot differentiate OBSERVER tools (legitimate
  observations) from GENERATOR tools (propositions masquerading as
  observations).
\item
  \textbf{Self-licensing is structural}: The problem is not that
  \(\text{LLM}_2\) is ``bad''. Even an ideal \(\text{LLM}_2\) produces
  propositions, not observations. Improving model quality does not
  eliminate type confusion.
\item
  \textbf{Channel \(\neq\) warrant}: The transport mechanism (tool call,
  API, memory injection) is causally relevant but epistemically
  irrelevant. Epistemic warrant must derive from content, not channel.
\end{enumerate}

\subsection{5. Architectural Patterns Where Laundering
Occurs}\label{architectural-patterns-where-laundering-occurs}

In this section, we enumerate not formally independent mechanisms, but
\textbf{recurring architectural patterns manifesting the same invariant
--- semantic laundering}, as defined in Definition 3. The presented
patterns differ in their manifestation level (instrumental,
interpretational, evaluative), but in all cases they realize the same
structural defect: epistemic status elevation without epistemically
relevant inference.

\subsubsection{5.1 Pattern 1: Tool Boundary
Laundering}\label{pattern-1-tool-boundary-laundering}

\textbf{Architectural pattern} (present in ReAct-style frameworks):

\begin{Shaded}
\begin{Highlighting}[]
\NormalTok{expert\_tool }\OperatorTok{=}\NormalTok{ LLMExpertTool()}
\NormalTok{result }\OperatorTok{=}\NormalTok{ expert\_tool.call(}\StringTok{"Diagnose patient symptoms"}\NormalTok{)}
\CommentTok{\# result: "Pneumonia, prescribe amoxicillin" (hypothesis, 0.65)}

\NormalTok{system.add\_observation(result)   }\CommentTok{\# ← accepted as observation}
\NormalTok{system.license }\OperatorTok{=}\NormalTok{ ASSERTIVE       }\CommentTok{\# ← now permits categorical }
                                 \CommentTok{\#   assertion}
\end{Highlighting}
\end{Shaded}

\textbf{Analysis}: LLM-generated hypothesis is included in observation
set \(O\) without epistemic typing. In terms of Definition 3 (semantic
laundering), this represents epistemic status elevation without
epistemically relevant inference.

\subsubsection{5.2 Pattern 2: Generation-Quality Warrant
Confusion}\label{pattern-2-generation-quality-warrant-confusion}

\textbf{Pattern}: Treating generation quality as epistemic warrant.

Under Definition 2, improvements in generation quality affect internal
coherence but do not constitute epistemically relevant inference unless
they introduce new observations or validated rules.

\textbf{Implicit assumption}: ``Expert'' LLM → better reasoning →
stronger warrant.

\textbf{Analysis}:

\begin{verbatim}
Expert LLM capabilities:
- More coherent argumentation
- Better approximation
- Higher text coherence

Expert LLM limitations:
- Cannot create observations
- Cannot confirm hypotheses
- Cannot escape proposition space P
\end{verbatim}

\textbf{Laundering mechanism}: Generation quality (argument structure,
coherence, persuasiveness) is conflated with warrant (grounding
quality). According to Definition 2 (epistemic relevance), generation
improvement is not epistemically relevant if observation set \(O\) has
not changed.

\subsubsection{5.3 Pattern 3: Architecturally-Mediated
Self-Licensing}\label{pattern-3-architecturally-mediated-self-licensing}

\textbf{Direct instantiation of Theorem 1}:

\begin{itemize}
\tightlist
\item
  Agent \(\text{LLM}_1\): Formulates question ``Should I do X?''
\item
  Expert \(\text{LLM}_2\) (via tool \(T\)): Returns ``Yes, do X''
  (hypothesis, \(P_2 \in P\))
\item
  System: Accepts \(P_2\) as observation (\(P_2 \in O\))
\item
  Admissibility decision: Uses \(P_2\) to permit ASSERTIVE statement by
  \(\text{LLM}_1\)
\end{itemize}

\textbf{Analysis}: This is exactly the circular justification mechanism
proven in Theorem 1. The agent validates itself through architectural
mediation.

\textbf{Critical observation}: At the code level, this circularity is
non-obvious (tool call appears as external interaction). It becomes
visible only when analyzing epistemic types.

\subsubsection{5.4 Pattern 4: Evaluator
Laundering}\label{pattern-4-evaluator-laundering}

\textbf{Observation}: The evaluator is subject to the same architectural
problem.

\textbf{Setup}: LLM-as-judge is used to evaluate agent output by
completeness or quality criteria.

\textbf{Problem}: When evaluator and agent share the same proposition
space (both are LLMs), evaluator interpretation variance manifests as
agent ``drift.''

\textbf{Corollary}: This is another instantiation of Theorem 1
(self-licensing). The evaluator's judgment is a proposition in the same
space as the agent's output, creating circular epistemic validation
under Definition 3 and Theorem 1.

\textbf{Connection to Corollary 1}: This pattern clearly shows why
LLM-as-judge is structurally unsound without epistemic typing. The
evaluative judgment and the evaluated proposition belong to the same
type, making validation circular if proposition types are not explicitly
distinguished.

\subsection{6. Implications for Safety-Critical
Deployment}\label{implications-for-safety-critical-deployment}

\subsubsection{6.1 Structural Impossibility of Compliance
Demonstration}\label{structural-impossibility-of-compliance-demonstration}

\textbf{We do not claim regulatory expertise.} We identify structural
impossibilities.

\textbf{Regulatory requirements} include:

\begin{itemize}
\tightlist
\item
  Traceability of decisions to evidence,
\item
  Demonstrable non-circularity of reasoning,
\item
  Failure mode analysis.
\end{itemize}

\textbf{In architectures with semantic laundering} (when Assumptions
1--3 are satisfied):

\begin{itemize}
\tightlist
\item
  The evidence chain breaks at the level of ``the tool said so'' ---
  what observations underlie the tool's conclusion?
\item
  Reasoning circularity is architecturally hidden --- LLM validates LLM
  through tool boundary,
\item
  Failure attribution collapses to opaque model behavior, eliminating
  causal traceability to observations.
\end{itemize}

\textbf{Structural impossibility}: These properties cannot be achieved
in architectures satisfying Assumptions 1--3, regardless of model
quality or training data.

\subsubsection{6.2 Auditability Failure}\label{auditability-failure}

\textbf{Post-incident analysis} in an architecture with semantic
laundering:

\begin{verbatim}
Question: "Why did the system make this decision?"
Answer: "Expert tool recommended it"

Question: "What observations grounded the tool output?"
Answer: "Patterns in model weights"

Question: "Can you show the justification chain?"
Answer: "No — it terminates in the tool"
\end{verbatim}

\textbf{Result}: The decision cannot be traced to observations. The
audit trail is structurally incomplete.

\subsubsection{6.3 Direction}\label{direction}

Avoiding this failure mode requires \textbf{explicit epistemic typing at
the architectural level} --- distinguishing tools that observe world
state from tools that generate propositions. A companion specification
proposes one concrete framework implementing this distinction
\citeproc{ref-romanchuk2026NormativeAdmissibilityFramework}{{[}11{]}}.

\subsection{7. Discussion}\label{discussion}

The result presented in this work is architectural, not empirical.
Theorem 1 follows directly from its premises and does not admit
statistical validation or performance benchmarking.

The purpose of architectural examples (§4) is exclusively to demonstrate
that these premises are realized in common agent runtimes under standard
design decisions. The work makes no claims about language model quality,
task execution efficiency, practical value, implementation quality, or
comparative merits of mentioned frameworks, nor about cross-domain
generalizability of presented observations.

\subsection{8. Related Problems and
Distinctions}\label{related-problems-and-distinctions}

\subsubsection{8.1 Distinction from Other Failure
Modes}\label{distinction-from-other-failure-modes}

\textbf{Hallucinations:}\\
Model generates factually incorrect content.\\
\textbf{vs.~Semantic laundering:} Content (correct or incorrect)
receives unjustified epistemic status.

\textbf{Prompt injection:}\\
Malicious input subverts agent behavior.\\
\textbf{vs.~Semantic laundering:} Architecture subverts its own
epistemic accounting.

\textbf{Reward hacking:}\\
Agent exploits evaluation metric.\\
\textbf{vs.~Semantic laundering:} Evaluation metric exploits
architectural type confusion.

The key distinction is that semantic laundering is not a model error,
optimization failure, or input vulnerability. It is a consequence of
systematically conflating causal and epistemic roles.

\subsubsection{8.2 Broader Implications}\label{broader-implications}

\textbf{For AI safety research}: Architectural epistemology is separable
from capability alignment questions. Improving model quality does not
eliminate type confusion or prevent circular justification.

\textbf{For practical deployment}: A specification like ``LLM-based
system'' is insufficient. Epistemic architecture (which components
observe the world and which generate propositions) must be specified
explicitly.

\textbf{For regulation and audit}: Compliance requirements cannot be met
without non-circular justification chains traceable to observations.
Architectures permitting semantic laundering structurally do not allow
building compliant chains.

\subsection{9. Related Work}\label{related-work}

\textbf{Epistemic logic}
\citeproc{ref-halpern2004ReasoningKnowledge}{{[}12{]}}: Formal systems
for reasoning about knowledge and belief. We apply these ideas to the
epistemic typing of AI agent architectures.

\textbf{Belief revision}
\citeproc{ref-gardenfors1988Knowledgefluxmodeling}{{[}13{]}}: Rational
knowledge updating. We address prevention of unwarranted belief
acquisition through architectural constraints.

\textbf{Tool use in LLMs}
\citeproc{ref-schick2023ToolformerLanguageModels}{{[}14{]}},
\citeproc{ref-patil2023GorillaLargeLanguage}{{[}15{]}}: Existing work
focuses on expanding model capabilities and API access. Tool outputs are
treated as authoritative signals by construction. We address a different
question: the epistemic status of tool outputs and how they should be
typed architecturally.

\textbf{LLM-as-judge}
\citeproc{ref-zheng2023JudgingLLMasaJudgeMTBench}{{[}7{]}},
\citeproc{ref-liu2023GEvalNLGEvaluation}{{[}8{]}}: Known problems
(position and length bias, self-preference). We identify structural
cause: with a common proposition space, evaluation becomes
self-licensing (Corollary 1). These effects align with analyses showing
fundamental epistemic fault lines between human judgment and LLM
generation, where plausibility may substitute for true justification
\citeproc{ref-quattrociocchi2025EpistemologicalFaultLines}{{[}16{]}}.

\textbf{AI safety and alignment}
\citeproc{ref-amodei2016ConcreteProblemsAI}{{[}17{]}},
\citeproc{ref-hendrycks2022UnsolvedProblemsML}{{[}18{]}}: Prior work has
articulated foundational safety challenges and open problems in ensuring
reliable and aligned behaviour of AI systems. We contribute an
architectural-epistemological perspective that highlights warrant
accounting as necessary for traceability and non-circular justification.

\subsection{10. Future Directions}\label{future-directions}

\subsubsection{10.1 Generalizability of the Architectural
Result}\label{generalizability-of-the-architectural-result}

This work is architectural in nature and not tied to a specific domain
or task type. Nevertheless, the question remains open as to which
classes of agent systems and runtimes manifest semantic laundering
patterns most clearly, and in which they may be partially mitigated
through additional constraints.

Further research could be directed toward systematic analysis of
architectural variations of agent runtimes and their epistemic
properties, without appeal to domain-specific scenarios.

\subsubsection{10.2 Multi-Agent
Architectures}\label{multi-agent-architectures}

Extending the analysis to multi-agent systems raises additional
architectural questions:

\begin{itemize}
\tightlist
\item
  How epistemic warrant propagates during inter-agent interaction,
\item
  Whether compositional preservation of epistemic typing is possible,
\item
  How circular justification manifests in hierarchical and cooperative
  agent structures.
\end{itemize}

This work is limited to analyzing architectures where agent interactions
reduce to proposition exchange in a common assumption space.

\subsubsection{10.3 Formal Type System}\label{formal-type-system}

A natural direction for further research of the presented analysis is
development of a formal epistemic type system, including:

\begin{itemize}
\tightlist
\item
  Explicit typing of tool results (e.g., OBSERVED vs INFERRED),
\item
  Verification of justification chains for non-circularity,
\item
  Automatic detection of architectural self-licensing patterns.
\end{itemize}

This work does not introduce such a type system; we only establish the
architectural necessity of epistemic distinction as a minimal condition
for eliminating semantic laundering.

\subsection{11. Conclusion}\label{conclusion}

In this work, we formalized \textbf{semantic laundering} --- an
architectural pattern where propositions acquire high epistemic status
by traversing superficially authoritative architectural boundaries
(tools, ``expert'' models) without epistemically relevant inference.

\subsubsection{Diagnosis}\label{diagnosis}

Semantic laundering constitutes an \textbf{architectural realization of
the Gettier problem}: propositions receive knowledge status without a
connection between justification and what makes them true. Unlike
classical Gettier cases, this effect is not accidental --- it is
architecturally determined and systematically reproduced when
Assumptions 1--3 are satisfied, corresponding to the dominant pattern of
contemporary agent frameworks.

\subsubsection{Why It Cannot Be Fixed}\label{why-it-cannot-be-fixed}

Scaling, model improvement, LLM-as-judge schemes, and evaluator
calibration cannot eliminate the problem. It exists at the type level,
not the generation level. Even an ideal LLM remains a GENERATOR --- it
produces propositions, not observations. Improving generation quality is
not epistemically relevant if the observation set has not changed.

\subsubsection{Architectural Response}\label{architectural-response}

We do not propose a solution to the Gettier problem in the philosophical
sense. Instead, we show that with explicit architectural distinction of
epistemic roles --- between observation, computation, and generation ---
the conditions necessary for Gettier situations to arise are
systematically excluded at the system level. The proposed approach does
not answer the question of what makes a belief knowledge. It solves an
engineering problem: defining admissible operations on propositions such
that circular justification and architecturally mediated self-licensing
are excluded.

In this mode, classical epistemological objections become irrelevant ---
the system does not realize the premises on which they are based.

\subsection*{References}\label{references}
\addcontentsline{toc}{subsection}{References}

\protect\phantomsection\label{refs}

\end{document}